\documentclass[twocolumn]{article}

\usepackage[
top    = 2.3cm,
bottom = 2.5cm,
left   = 2.0cm,
right  = 2.0cm]{geometry}

\usepackage{times}

\usepackage{latexsym} 

\usepackage{natbib}

\usepackage{graphicx}
\usepackage{url}
\usepackage{amsmath,amssymb}
\usepackage{xspace}

\newcommand{\ours}{{EigenGP}\xspace}
\newcommand{\egp}{{EigenGP}\xspace}
\newcommand{\GP}{Gaussian process\xspace}

\newcommand{\eg}{{\em e.g.}\xspace}
\newcommand{\ie}{i.e.\xspace}

\renewcommand{\b}{{\bf b}}
\renewcommand{\d}{{\rm d}}  % for derivatives
\newcommand{\f}{{\bf f}}
\renewcommand{\k}{{\bf k}}
\newcommand{\m}{{\bf m}}
\renewcommand{\u}{{\bf u}}
\newcommand{\w}{{\bf w}}
\newcommand{\x}{{\bf x}}
\newcommand{\y}{{\bf y}}
\newcommand{\B}{{\bf B}}
\newcommand{\C}{{\bf C}}
\newcommand{\I}{{\bf I}}
\newcommand{\K}{{\bf K}}
\renewcommand{\P}{{\bf P}}
\newcommand{\Q}{{\bf Q}}
\newcommand{\R}{{\bf R}}
\renewcommand{\S}{{\bf S}}
\newcommand{\X}{{\bf X}}

\newcommand{\N}{{\mathcal N}}  % for normal density

\newcommand{\balpha}{\boldsymbol{\alpha}}
\newcommand{\btheta}{\boldsymbol{\theta}}
\newcommand{\Beta}{\boldsymbol{\eta}}
\newcommand{\bPhi}{\mathbf{\Phi}}

\newcommand{\1}{{\bf 1}}
\newcommand{\0}{{\bf 0}}

\newcommand{\Tr}{^{\rm T}}
\newcommand{\diag}{{\rm diag}}
\newcommand{\tr}{{\rm tr}}
\renewcommand{\vec}{{\rm vec}}
\renewcommand{\det}[1]{\left| #1 \right|}

\newcommand{\fx}{ \f_{\x} \xspace}
\newcommand{\tK}{\tilde{\K}\xspace}

\newcommand{\Kbb}{{\K_{BB}}}
\newcommand{\Kbx}{{\K_{BX}}}
\newcommand{\Kxb}{{\K_{XB}}}

\title{EigenGP: Gaussian Process Models with Adaptive Eigenfunctions}
\date{}
\author{
    Hao Peng\\
\small Departments of  Computer Science\\
\small Purdue University\\
\small West Lafayette, IN 47906\\
\small pengh@purdue.edu\\
  \and
    Yuan Qi\\
\small Departments of CS and Statistics\\
\small Purdue University\\
\small West Lafayette, IN 47906\\
\small alanqi@purdue.edu\\
}
\begin{document}

\maketitle

\begin{abstract}
 Gaussian processes (GPs) provide a nonparametric
  representation of functions. However, classical GP inference suffers from high computational cost for big data. 
  In this paper, we propose a new Bayesian approach, \ours, that 
  learns both basis dictionary elements---eigenfunctions of a GP prior---and prior precisions in a sparse finite model. It is well known that, among all orthogonal basis functions, % including trigonometric basis functions, 
  eigenfunctions can provide the most compact representation. Unlike other sparse Bayesian finite models where the basis function has a fixed form, our eigenfunctions live in a reproducing kernel Hilbert space as a finite linear combination of kernel functions. 
  We learn the dictionary elements---eigenfunctions---and the prior precisions over these elements as well as all the other hyperparameters from data by maximizing the model marginal likelihood.
  We explore computational linear algebra to simplify the gradient computation significantly.
  Our experimental results demonstrate improved predictive performance of
  \egp over alternative sparse GP methods as well as relevance vector machines.
\end{abstract}

\section{Introduction}
Gaussian processes (GPs)  are powerful nonparametric Bayes\-ian models with numerous applications in  machine learning and statistics.
GP inference, however, is costly. Training the exact GP regression model with $N$ samples is expensive: it takes an $O(N^2)$ space cost %to save the whole covariance matrix 
and an $O(N^3)$ time cost. % for inverting this matrix. For big data with large $N$, the cost of GP inference will be prohibitively high. 
To address this issue, a variety of approximate sparse GP inference approaches
have been developed \citep{Williams01,Csato2002,Snelson06sparsegaussian,Gredilla10,WillamsBarber98,Titsias09,Qi:SparsePosteriorGaussianProc,Higdon2002,CressieAndJohannesson08}---for example, using the Nystr\"{o}m method to approximate covariance matrices \citep{Williams01}, optimizing a variational bound on the marginal likelihood~\citep{Titsias09} or grounding  the GP on a
small set of  (blurred) basis  points \citep{Snelson06sparsegaussian,Qi:SparsePosteriorGaussianProc}.
An elegant unifying view for various sparse GP regression models is given by \citet{Candela05Rasmussen}.

Among all sparse GP regression methods, a state-of-the-art approach is to represent a function as a sparse finite linear combination of pairs of trigonometric basis functions, 
 a sine and a cosine for each spectral point; thus this approach is called sparse spectrum Gaussian process (SSGP) \citep{Gredilla10}. 
SSGP integrates out both weights and phases of the  trigonometric functions and learns all hyperparameters of the model (frequencies and amplitudes) by maximizing the marginal likelihood. Using global trigonometric functions as basis functions,  SSGP has the capability of approximating any stationary Gaussian process model and been shown to outperform alternative sparse GP methods---including fully independent training conditional (FITC) approximation \citep{Snelson06sparsegaussian}---on benchmark datasets. 
Another popular sparse Bayesian finite linear model is the relevance vector machine (RVM) \citep{Tipping00,Faul01analysisof}. It uses kernel expansions over training samples as basis functions and selects the basis functions by automatic relevance determination \citep{MacKay92,Faul01analysisof}.
%based on expectation maximization \citep{MacKay92} or fast fixed point updates \citep{Faul01analysisof}.

In this paper, we propose a new sparse Bayesian approach, \ours, that 
learns both functional dictionary elements---eigenfunctions---and prior precisions in a finite linear model representation of a GP. It is well known that, among all orthogonal basis functions, % including trigonometric basis functions, 
eigenfunctions provide the most compact representation. Unlike SSGP or RVMs where the basis function has a fixed form, our eigenfunctions live in a reproducing kernel Hilbert space (RKHS) as a finite linear combination of kernel functions with their weights learned from data. 
We further marginalize out weights over eigenfunctions and estimate all hyperparameters---including basis points for eigenfunctions, lengthscales, and precision of the weight prior---by maximizing the model marginal likelihood (also known as evidence). To do so, we explore computational linear algebra and greatly simplify the gradient computation for optimization (thus our optimization method is totally different from RVM optimization methods). As a result of this optimization, our eigenfunctions are data dependent and make \ours capable of accurately modeling nonstationary data.
Furthermore, by adding an additional kernel term in our model,  we can turn the finite model into an infinite model to model the prediction uncertainty better---that is, it can give nonzero prediction variance when a test sample is far from the training samples. 

% From the dictionary learning perspective, \ours integrates out weights over dictionary elements to avoid overfitting and learns dictionary elements---eigenfunctions---from data.

\ours is computationally efficient. It takes $O(NM)$ space and $O(NM^2)$ time for training on with $M$ basis functions, which is same as SSGP and more efficient than RVMs (as RVMs learn weights over $N$, not $M$, basis functions.). 
Similar to FITC and SSGP, \ours focuses on predictive accuracy at low computational
cost, rather than on faithfully converging towards the full GP as the number of basis functions grows. (For the latter case, please see the approach \citep{Yan10} that explicitly minimizes the KL divergence between exact and approximate GP posterior processes.)

The rest of the paper is organized as follows. Section 2 describes the background of GPs. Section 3 presents the \ours model and an illustrative example. Section 4 outlines the marginal likelihood maximization for learning dictionary elements and the other hyperparameters. In Section 5, we discuss related work. Section 6 shows regression results on multiple benchmark regression datasets, demonstrating improved performance of \ours over Nystr\"om~\citep{Williams01}, RVM, FITC, and SSGP.

%%--------------------------------------------------------------------------------------------------
\section{Background of Gaussian Processes}
We denote $N$ independent and identically distributed samples as
$\mathcal{D}=\{(\x_{1},y_{1}),\ldots,(\x_{n},y_{n})\}_N$, where
$\x_i$ is a $D$ dimensional input (\ie, explanatory variables) and $y_i$ is a scalar output (\ie, a response), which we assume is
the noisy realization of a latent function $f$ at $\x_i$.

A \GP places a prior distribution over the latent function $f$. 
Its projection $\fx$ at  $\{\x_i\}_{i=1}^N$ defines a joint Gaussian distribution
$p(\fx) = \N(\f|\m^0, \K)$,
where, without any prior preference, the mean $\m^0$ are set to $\0$ and the covariance function $k(\x_i,\x_j)\equiv\K(\x_i,\x_j)$ encodes the
prior notion of smoothness. A popular choice 
is the anisotropic squared exponential covariance function:
%\begin{eqnarray}\label{eq:rbf}
$k(\x, \x') = a_0 \exp\left(-(\x-\x')\Tr \diag(\Beta) (\x-\x')\right),$
%\end{eqnarray}
where the hyperparameters include the signal variance  $a_0$ and the lengthscales $\Beta=\{\eta_d\}_{d=1}^D$, controlling how fast the covariance decays with the distance between inputs.
Using this covariance function, we can prune input dimensions by shrinking the corresponding lengthscales based on the data (when $\eta_d=0$, the $d$-th dimension becomes totally irrelevant to the covariance function value). This pruning is known as Automatic Relevance Determination (ARD) and therefore this covariance is also called the ARD squared exponential. 
Note that the covariance function value remains the same when
$(\x'-\x)$ is the same -- regardless where
$\x'$ and $\x$ are. This thus leads to a {\em stationary} GP model. For nonstationary data, however, a stationary GP model is a misfit. 
Although nonstationary GP models have been developed and applied to real world applications,
they are often limited to  low-dimensional problems, such as applications in spatial statistics \citep{PacSch03}.
Constructing general nonstationary GP models remains a challenging task. 

For regression, we use a Gaussian likelihood function $p(y_i|f) = \N(y_i|f(\x_i),\sigma^2)$,
%\begin{align}\label{reglik}
%p(y_i|f) = \N(y_i|f(\x_i),\sigma^2),
%\end{align}
where $\sigma^2$ is the variance of the observation noise. 
Given the \GP prior over $f$ and the data likelihood, the exact posterior process is
%\begin{align}\label{eq:exact}
$p(f|\mathcal{D},\y) \propto GP(f|0,k) \prod_{i=1}^{N} p(y_i|f).$
%\end{align}
Although the posterior process for GP regression has an analytical form, we need to store and 
invert an $N$ by $N$ matrix, which has the computational complexity $O(N^3)$, rendering GP unfeasible for big data analytics.	

%%-------------------------------------------------------------------------------------------------
\section{Model of \egp}\label{sec:model}
To enable fast inference and obtain a nonstationary covariance function, our new model \egp projects the GP prior in an eigensubspace. Specifically, we set the latent function %$f$
\begin{align}\label{eq:klexpan}
f(\x)  =  \sum_{j=1}^M \alpha_j \phi^j(\x)
\end{align}
where $M\ll N$ and $\{\phi^j(\x)\}$ are eigenfunctions of the GP prior.
We assign a Gaussian prior over $\balpha=[\alpha_1,\ldots,\alpha_M]$, $\balpha\sim\N(\0,\diag(\w)),$
%\begin{align}\label{eq:prior_theta}
%\balpha\sim\N(\0,\diag(\w)),
%\end{align}
so that $f$ follows a GP prior with zero mean and the following covariance function
\begin{align}\label{eq:ktilde}
\tilde{k}(\x,\x') = \sum_{j=1}^M w_j \phi^j(\x) \phi^j(\x').
\end{align}

To compute the eigenfunctions $\{\phi^j(\x)\}$, we can use the Galerkin projection to approximate them by Hermite polynomials \citep{Marzouk09}. For high dimensional problems, however, this approach requires a tensor product of univariate Hermite polynomials that dramatically increases the number of parameters.

To avoid this problem, we apply the Nystr\"om  method \citep{Williams01} that allows us to obtain an approximation to the eigenfunctions in a high dimensional space efficiently.
Specifically, given inducing inputs (\ie basis points)
$\B = [\b_1,\ldots,\b_M]$, 
we replace
\begin{align} \label{eq:mercer}
\int k(\x,\x') \phi^j(\x) p(\x)\d\x =  \lambda_j \phi^j(\x')
\end{align}
by its Monte Carlo approximation
\begin{align}
\frac{1}{M}\sum_{i=1}^M  k(\x,\b_i) \phi^j(\b_i) \approx \lambda_j  \phi^j(\x)
\end{align}
Then, by evaluating this equation at $\B$ so that we can estimate the values of $\phi^j(\b_i)$ and $\lambda_j$, we obtain the $j$-th eigenfunction $\phi^j(\x)$ as follows
\begin{align} \label{eq:eigf}
%\phi^j(\x) = \frac{\sqrt{M}}{\lambda_j^{(M)}}\k(\x)  \tu_j = {\k(\x) \u_j}
\phi^j(\x) = \frac{\sqrt{M}}{\lambda_j^{(M)}}\k(\x)  \u_j^{(M)} = {\k(\x) \u_j}
\end{align}
where $\k(\x) \triangleq [k(\x,\b_1),\ldots, k(\x,\b_M)]$, 
$\lambda_j^{(M)}$ and $\u_j^{(M)}$ are the $j$-th eigenvalue and eigenvector of the covariance function evaluated at $\B$,
and %$\u_j = \frac{\sqrt{M}}{\lambda_j^{(M)}} \u_j^{(M)}$. 
$\u_j = \sqrt{M}\u_j^{(M)}/\lambda_j^{(M)}$. 
Note that, for simplicity, we have chosen the number of the inducing inputs to be the same as the number of the eigenfunctions; in practice, we can use more inducing inputs while computing only the top $M$ eigenvectors.
As shown in  \eqref{eq:eigf}, our eigenfunction lives in a RKHS as a linear combination of the kernel functions evaluated at $\B$ with weights $\u_j$.

Inserting \eqref{eq:eigf} into \eqref{eq:klexpan} we obtain
\begin{eqnarray}
f(\x)=\sum_{j=1}^M \alpha_j\sum_{i=1}^M u_{ij}k(\x,\b_i)% \nonumber
%\label{eq:rvm-sim}
\end{eqnarray}
This equation reveals a two-layer structure of \egp.
The first layer linearly combines multiple kernel functions to generate each eigenfunction $\phi^j$. The second layer takes
these eigenfunctions as the basis functions to generate the function value $f$. Note that $f$ is a Bayesian linear combination of  $\{\phi^j\}$ where the weights $\balpha$ are integrated out to avoid overfitting.  %Thus \egp can be viewed as a two-layer \comment{\em deep}  Bayesian kernel machine.  

All the model hyperparameters are learned from data. 
Specifically, for the first layer, 
 to learn the eigenfunctions $\{\phi_i\}$,  we estimate
the inducing inputs $\B$ and  the kernel hyperparameters (such as lengthscales $\Beta$ for the ARD kernel) by maximizing the model marginal likelihood. For the second layer, we marginalize out $\balpha$ to avoid overfitting and maximize the model marginal likelihood to learn the hyperparameter $\w$ of the prior. %in \eqref{eq:prior_theta}.

With the estimated hyperparameters, the prior over $f$ is nonstationary because its covariance function in \eqref{eq:ktilde} varies at different regions of $\x$. This comes at no surprise since the eigenfunctions are tied with $p(\x)$ in \eqref{eq:mercer}. This nonstationarity reflects the fact that our model is adaptive to the distribution of the explanatory variables $\x$.

Note that to recover the full uncertainty captured by the kernel function $k$, we can add the following term into the kernel function of \egp:
\begin{align} \label{eq:dif}
\delta(\x-\x') ( k(\x,\x') - \tilde{k}(\x,\x'))
\end{align}
where  $\delta(a)=1$ if and only if $a=0$.
Compared to the original \egp model, which has a finite degree of freedom, this modified model has the infinite number of basis functions (assuming $k$ has an infinite number of basis functions as the ARD kernel). Thus, this model can accurately model the uncertainty of a test point even when it is far from the training set.  
We derive the optimization updates  of all the hyperparameters for both the original and modified \egp models. But according to our experiments, the modified model does not improve the prediction accuracy over the original \egp  (it even reduces the accuracy sometimes.). Therefore, 
we will focus on the original \egp model in our presentation for its simplicity.  Before we present details about hyperparameter optimization, let us first look at an  illustrative example on the effect of hyperparameter optimization, in particular, the optimization of the inducing inputs $\B$.

\subsection{Illustrative Example}

\begin{figure}
\begin{center}
\includegraphics[width=0.9\linewidth]{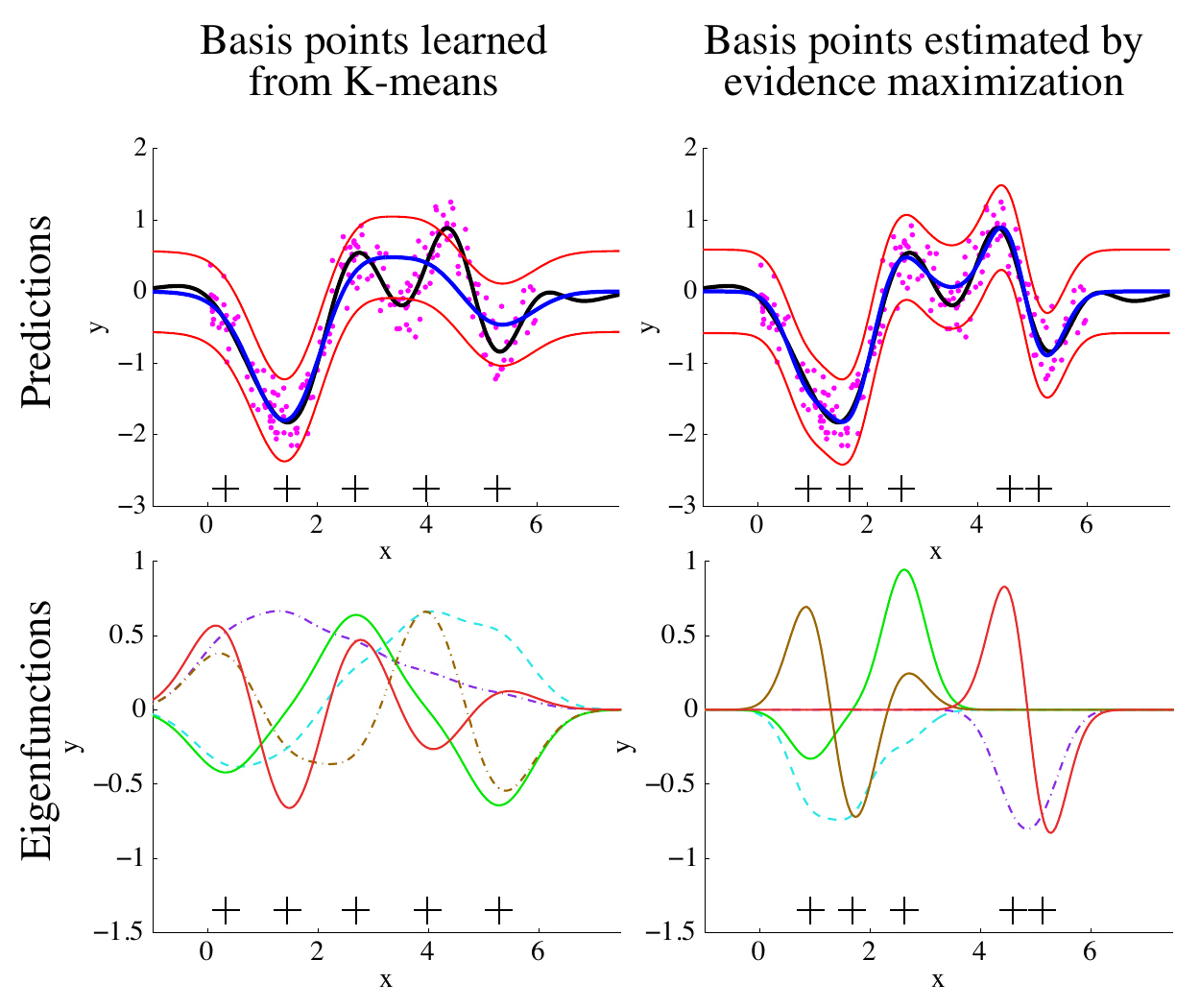}
\end{center}
\caption{Illustration of the effect of optimizing basis points (\ie, inducing inputs). In the first row, the pink dots represent data points, the blue and red solid curves correspond to 
the predictive mean of \egp and $\pm$ two standard deviations around the mean, the black curve corresponds to the predictive mean of the full GP, and the black crosses denote the basis points. In the second row, the curves in various colors represent the five eigenfunctions of \egp. 
} \label{fig:eigfun}
\end{figure}

For this example, we consider a toy dataset used by the FITC algorithm
\citep{Snelson06sparsegaussian}.
It contains 200 one-dimensional training samples.
We use 5 basis points ($M=5$) and choose the ARD kernel. 
We then compare the basis functions $\{\phi^j\}$ and the corresponding predictive distributions in two cases. For the first case, we use the kernel width $\eta$ learned from the full GP model as the kernel width for \egp,  and
apply K-means to set the basis points $\B$ as cluster centers.
The idea of using K-means to set the basis points has been suggested by \citet{ZhangNystrom08} to minimize an error bound for the Nystr\"{o}m  approximation. For the second case, we optimize $\eta$, $\B$ and $\w$ by
maximizing the marginal likelihood of \ours.

The results are shown in Figure \ref{fig:eigfun}. The first row demonstrates that, by optimizing the hyperparameters, \egp achieves the predictions very close to what the full GP achieves---but using only 5 basis functions. In contrast, when the basis points are set to the cluster centers by K-means, \egp leads to the prediction significantly different from that of the full GP and fails to capture the data trend, in particular, for $x\in (3,5)$. 
The second row of Figure \ref{fig:eigfun} shows that  K-means sets the basis points almost evenly spaced on $y$, and accordingly the five eigenfunctions are  
 smooth {\em global} basis functions whose shapes are not directly linked to the function they fit. Evidence maximization, by contrast, sets the  basis points unevenly spaced to generate the
 basis functions whose shapes are more {\em localized} and  adaptive to the function they fit; for example, the eigenfunction represented by the red curve well matches the data on the right. 

\section{Learning Hyperparameters}
In this section we describe how we optimize all the hyperparameters, denoted by  $\btheta$, which include $\B$ in the covariance function \eqref{eq:ktilde}, and all the kernel hyperparameters (e.g., $a_0$ and $\Beta$). 
To optimize $\btheta$, we maximize the  marginal likelihood  (\ie evidence) based on a conjugate Newton method\footnote{We use the code
 % in the GPML package: 
 from \url{http://www.gaussianprocess.org/gpml/code/matlab/doc}}. 
We explore two strategies for evidence maximization. The first one is sequential optimization, which first 
fixes $\w$ while updating all the other hyperparameters, and then optimizes $\w$ while fixing the other hyperparameters. The second strategy is to optimize all the hyperparameters jointly. Here we 
skip the details for the more complicated joint optimization 
and describe the key gradient formula for sequential optimization. % In practice, the sequential optimization is computationally more efficient and less prone to bad local optima than the joint optimization.
The key computation in our optimization is for the log marginal likelihood
and its gradient: % in the vector or matrix form:
\small
\begin{eqnarray}
\ln p(\y|\btheta)  = & -\frac{1}{2}\ln \det{\C_N}-\frac{1}{2}\y\Tr\C_N^{-1}\y -\frac{N}{2}\ln(2\pi), \label{eq:logmarglik} \\
 \d\ln p(\y|\btheta) = & -\frac{1}{2}\left[\tr(\C_N^{-1}\d{\C_N})- \tr(\C_N^{-1}\y\y\Tr\C_N^{-1}\d\C_N) \right] \label{eq:diffmarg}
\end{eqnarray}
\normalsize
where 
$\C_N =\tK + \sigma^2 \I$,
$\tK = \bPhi \diag(\w) \bPhi\Tr $, 
and $\bPhi = \{\phi^m(\x_n)\}$ is an $N$ by $M$ matrix.
Because the rank of $\tK$ is $M\ll N$, we can compute 
$\ln\det{\C_N}$ and $\C_N^{-1}$ efficiently with the cost of $O(M^2N)$ via the matrix inversion and determinant lemmas. Even with the use of the matrix inverse lemma for the low-rank computation, a naive calculation would be very costly. We apply identities from computational linear algebra \citep{Minka01oldand,Leeuw07} to simplify the needed computation dramatically.  

%\subsection{Derivatives with Respect to $\B$, $\Beta$, $a_0$ and $\sigma^2$} 
To compute the derivative with respect to $\B$, we first notice that, when $\w$ is fixed, we have
%\small
\begin{align}\label{eq:seq_CN}
\C_N =\tK + \sigma^2 \I = 
\Kxb \Kbb^{-1}\Kbx +\sigma^2\I 
\end{align}
%\normalsize
where $\Kxb$ is the cross-covariance matrix between the training data $\X$ and the inducing inputs $\B$, and $\Kbb$ is the covariance matrix on $\B$. 

For the ARD squared exponential kernel,
utilizing the following identities,
$\tr(\P\Tr\Q)  = \vec(\P)\Tr\vec(\Q)$ and
$\vec(\P\circ \Q) = \diag(\vec(\P)) \vec(\Q)$,
where $\vec(\cdot)$ vectorizes a matrix into a column vector, and $\circ$ represents the Hadamard product, 
we can derive the derivative of the first trace term in \eqref{eq:diffmarg}%\eqref{eq:eq7}:
\small
\begin{align}
& \frac{\tr(\C_N^{-1}\d\C_N)}{\d\B} = 4\R \X\Tr\diag(\Beta) - 4(\R \1 \1\Tr)\circ(\B\Tr\diag(\Beta))  \nonumber \\
& \quad\quad\quad\quad\quad -4\S\B\Tr\diag(\Beta) +4(\S \1 \1\Tr)\circ(\B\Tr\diag(\Beta))
\end{align}
\normalsize
where $\1$ is a column vector of all ones, and
%\small
\begin{eqnarray}
\R & = & (\Kbb^{-1}\Kbx\C_N^{-1})\circ\Kbx \label{eq:R} \\
\S & = & (\Kbb^{-1}\Kbx\C_N^{-1}\Kxb\Kbb^{-1})\circ\Kbb \label{eq:S}
\end{eqnarray}
%\normalsize
Note that we can compute $\Kbx\C_N^{-1}$ efficiently via low-rank updates. Also, $\R \1 \1\Tr$ ($\S \1 \1\Tr$) can be implemented efficiently by first summing over the columns of $\S$ ($\R$) and then copying it multiple times---without any multiplication operation.
To obtain $\frac{\tr(\C_N^{-1}\y\y\Tr\C_N^{-1}\d\C_N)}{\d\B}$, we simply replace $\C_N^{-1}$ in  
(\ref{eq:R}) and (\ref{eq:S}) by $\C_N^{-1}\y\y\Tr\C_N^{-1}$.

Using similar derivations, we can obtain the derivatives with respect to the lengthscale $\Beta$, $a_0$ and $\sigma^2$ respectively.

%\subsection{Derivative with Respect to $\w$}
To compute the derivative with respect to  $\w$, we can use the formula
$\tr(\P \diag(\w)\Q) = \1\Tr(\Q\Tr\circ\P) \w$ to obtain the two trace terms in \eqref{eq:diffmarg} as follows:
\small 
\begin{align}
 \frac{\tr(\C_N^{-1}\d\C_N)}{\d\w}  & = 
\1\Tr (\bPhi \circ (\C_N^{-1}\bPhi)) \\
 \frac{\tr(\C_N^{-1}\y\y\Tr\C_N^{-1}\d\C_N)}{\d\w} & = 
\1\Tr (\bPhi \circ (\C_N^{-1}\y\y\Tr\C_N^{-1}  \bPhi)) 
\end{align}
\normalsize

%\subsection{Computational Complexity} 
For either sequential or joint optimization, the overall computational complexity is $O(\textrm{max}(M^2,D) N)$ where $D$ is the data dimension.

% Given the gradient formula, we apply a conjugate Newton method to learn all the hyperparameters.

\section{Related Work}\label{sec:relate}
Our work is closely related to the seminal work by~\citep{Williams01}, but they differ in multiple aspects. First, we define a valid probabilistic model based on an eigen-decomposition of the GP prior. By contrast, the previous approach \citep{Williams01}
aims at a low-rank approximation to the finite covariance/kernel matrix used in GP training---from a numerical approximation perspective---and its predictive distribution is not well-formed in a probabilistic framework (\eg, it may give a negative variance of the predictive distribution.). 
 Second, while the Nystr\"om method simply uses the first few eigenvectors, we maximize the model marginal likelihood to adjust their weights in the covariance function.
Third, exploring the clustering property of the eigenfunctions of the Gaussian kernel, our approach can conduct semi-supervised learning, while the previous one cannot. The semi-supervised learning capability of \egp is investigated in another paper of us.
Fourth, the Nystr\"{o}m method lacks a principled way to learn model hyperparameters including the kernel width and the basis points while \ours does not.
% Fourth, while the Nystr\"{o}m method can well approximate a stationary GP given a large number of basis points, the estimated basis points and weights in EigenGP's  covariance function make it suitable for nonstationary data.

Our work is also related to methods that use kernel principle component analysis (PCA) to speed up kernel machines
 \citep{Hoegaerts05}. 
However, for these methods it can be difficult---if not impossible---to learn important hyperparameters including kernel width for each dimension and inducing inputs (not a subset of the training samples). By contrast, \egp learns all these hyperparameters from data based on gradients of the model marginal likelihood.

\section{Experimental Results}
In this section, we compare \ours \footnote{The implementation is available at: \url{https://github.com/hao-peng/EigenGP}}
and alternative methods on synthetic and real benchmark datasets. 
The alternative methods include the sparse GP methods---FITC, SSGP, and the Nystr\"{o}m method---as well as RVMs. We implemented the Nystr\"{o}m method ourselves and downloaded the software implementations for the other methods from their authors' websites. For RVMs, we used the fast fixed point algorithm \citep{Faul01analysisof}.
%FITC, \url{http://www.gatsby.ucl.ac.uk/~snelson/SPGP_dist.zip}; SSGP, \url{http://www.tsc.uc3m.es/~miguel/code/ssgpr_code.zip};  and RVM, \url{http://www.vectoranomaly.com/downloads/SB2_Release_200.zip}.
We used the ARD kernel for all the methods except RVMs (since they do not estimate the lengthscales in this kernel) and optimized all the hyperparameters via evidence maximization. For RVMs, we chose the squared exponential kernel with the same lengthscale for all the dimensions and applied a 10-fold cross-validation on the training data to select the lengthscale. 
On large real data, we used the values of $\Beta$, $a_0$, and $\sigma^2$ learned from the full GP on a subset that was 1/10 of the training data to initialize all the methods except RVMs.
For the rest configurations, we used the default setting of the downloaded software packages.
For our own model, we denote the versions with sequential and joint optimization as \ours and EigenGP*, respectively.
To evaluate the test performance of each method, we measure the Normalized Mean Square Error (NMSE) and the Mean Negative Log Probability (MNLP), defined as:
\begin{eqnarray}
\text{NMSE} = & {\sum_i (y_{i}-\mu_{i})^2}/{\sum_i(\mu_{i}-\bar{y})^2} \\
\text{MNLP} = & \frac{1}{2N} \sum_i [ (\frac{y_{i}-\mu_{i}}{\sigma_{i}})^2+\ln\sigma_{i}^2+\ln2\pi ]
\end{eqnarray}
where $y_{i}$, $\mu_{i}$ and $\sigma_{i}^2$ are the response value, the predictive mean and  variance for the $i$-th test point respectively, and 
$\bar{y}$ is the average response value of the training data.

\subsection{Approximation Quality on Synthetic Data}\label{sec:uncertainty-comp-res}
\begin{figure}[th]
\begin{center}
\includegraphics[width=0.9\linewidth]{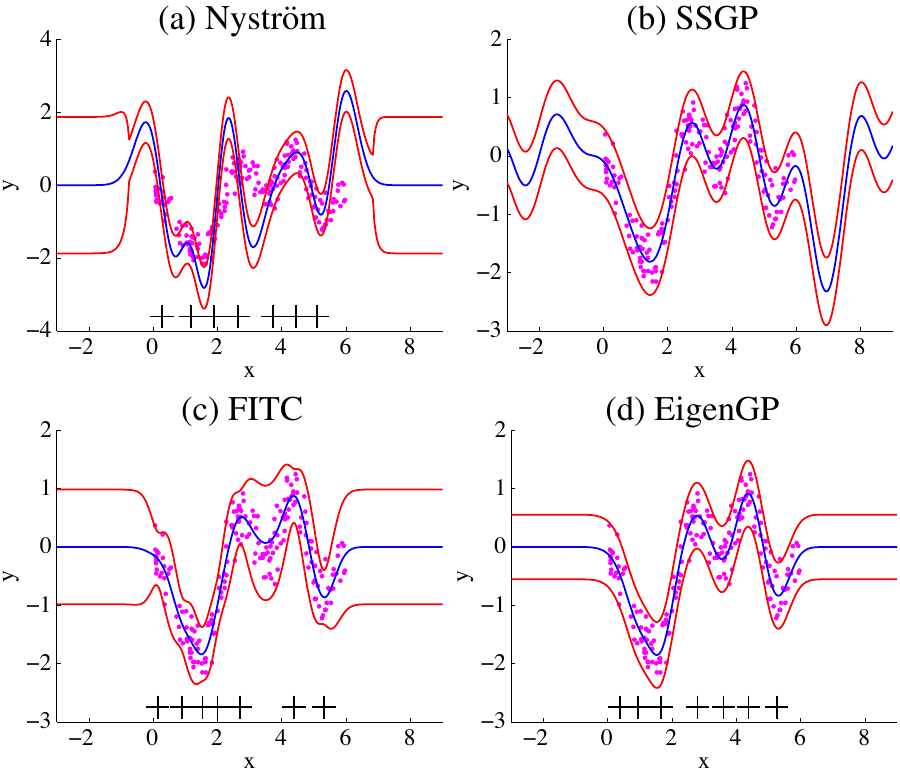}
\end{center}
\caption{Predictions of four sparse GP methods. Pink dots represent training data; blue curves are predictive means; red curves are two standard deviations above and below the mean curves, and the black crosses indicate the inducing inputs.} \label{fig:data-preds}
\end{figure}

As in Section 3.1, we use the synthetic data from the FITC paper for the comparative study. To let all the methods have the  same computational complexity, we set the number of inducing inputs $M=7$.
%$M$ is intentionally set to be less than $10$ that was used in the FITC paper to illustrate the case when the number of basis functions is very small.
The results are summarized in Figure \ref{fig:data-preds}.
For the Nystr\"{o}m method, we used the kernel width learned from the full GP and applied K-means to choose the basis locations \citep{ZhangNystrom08}. 
Figure \ref{fig:data-preds}(a) shows that it does not fit well.
Figure \ref{fig:data-preds}(b) demonstrates that the prediction of SSGP
oscillates outside the range of the training samples, probably due to the fact that the
sinusoidal components are {\em global} and span the whole data range (increasing the number of basis functions would improve SSGP's predictive performance, but increase the computational cost.). As shown by Figure \ref{fig:data-preds}(c), %although FITC provides better predictive uncertainty in the region far from the training data, it 
FITC fails to capture the turns of the data accurately for $x$ near $4$ while \ours can.
%. By contrast,  \ours better models the data than the other methods around the training data. 

Using the full GP predictive mean as the label for $x\in[-1, 7]$ (we do not have the true $y$ values in the test data), we compute the NMSE and MNLP of all the methods. The average results from 10 runs 
% % each with different random seeds,
 are reported in Table \ref{table:syn_nmse} (dataset 1). We have the results from two versions of the Nystr\"{o}m method. For the first version,
 the kernel width is learned from the full GP and the basis locations are chosen by K-means as before; for the second version, denoted as Nystr\"{o}m*, its hyperparameters are learned by evidence maximization. Note that the evidence maximization algorithm for the Nystr\"{o}m approximation is novel too---developed by us for the comparative analysis. 
Table \ref{table:syn_nmse} shows that both \ours and EigenGP* approximate the mean of the full GP model more accurately than the other methods, in particular, several orders of magnitude better than the Nystr\"{o}m method.
%We also compute the values of the KL divergence between the predictive distributions of the full GP and the sparse GPs and report them in Appendix. 
\begin{table}[t]
\caption{NMSE and MNLP on synthetic data } \label{table:syn_nmse}
\begin{center}
NMSE\\
\begin{tabular}{c|c|c}
% M = 15 seed=1
\hline
Method & dataset 1 & dataset 2\\
\hline 
\hline
Nystr\"{o}m & $39\pm 18$ & $1526\pm  769$  \\
\hline 
Nystr\"{o}m*  & $2.41 \pm 0.53$ & $2721\pm  370$  \\
\hline
FITC & $0.02 \pm 0.005$ & $  0.50\pm 0.04$   \\
\hline
SSGP & $ 0.54\pm  0.01$ & $ 0.22\pm0.05$ \\
\hline
EigenGP &  $  0.006 \pm 0.001$ & $ 0.06\pm  0.02$  \\
\hline
EigenGP$^*$  &  $0.009\pm0.002$  & $0.06\pm 0.02$  \\
%\hline
% EigenGP$^+$  &  $0.035\pm0.005$ & $0.39\pm0.01$ \\
%\hline
%FullGP& 0.03202 \\
\hline
\end{tabular} \\
\begin{tabular}{c}
\\
MNLP
\end{tabular}\\
\begin{tabular}{c|c|c}
% M = 15 seed=1
\hline
Method & dataset 1 & dataset 2\\
\hline 
\hline
Nystr\"{o}m & $645 \pm 56$ & $ 2561\pm 1617$  \\
\hline 
Nystr\"{o}m*  & $7.39 \pm  1.66$ & $40\pm5$  \\
\hline
FITC & $ -0.07 \pm  0.01$ & $0.88\pm0.05$   \\
\hline
SSGP & $  1.22\pm  0.03$ & $ 0.87\pm  0.09$ \\
\hline
EigenGP &  $ -0.33 \pm 0.00$ & $ 0.40\pm 0.07$  \\
\hline
EigenGP$^*$ & $-0.31\pm 0.01$  & $ 0.44\pm0.07$  \\
%\hline
% EigenGP$^+$  &  $0.035\pm0.005$ & $0.39\pm0.01$ \\
%\hline
%FullGP& 0.03202 \\
\hline
\end{tabular}
\end{center}
\end{table}

Furthermore, we add the difference term \eqref{eq:dif} into the kernel function and denote this version of our algorithm as EigenGP$^+$. It gives better predictive variance when far from the training data but its predictive mean is slightly worse than the version without this term \eqref{eq:dif}; the NMSE and MNLP of EigenGP$^+$ are $0.014\pm0.001$ and $-0.081\pm 0.004$. Thus, on the other datasets, we only use the versions  without this term (\ours and EigenGP$^*$) for their simplicity and effectiveness. 
We also examine the performance of all these methods with a higher computational complexity. Specifically, we set $M=10$.
Again, both  versions of the Nystr\"{o}m method  give poor predictive distributions. And SSGP still leads to extra wavy patterns outside the training data. FITC, \ours and  EigenGP$^+$ give good predictions. Again, EigenGP$^+$  gives better predictive variance when far from the training data %(Figure 1f in Appendix)
, but with a similar predictive mean as \ours.

Finally, we compare the RVM with EigenGP* on this dataset. While the RVM gives NMSE $=0.048$ in 2.0 seconds, EigenGP* achieves NMSE $=0.039 \pm 0.017$ in $0.33\pm 0.04$ second with $M=30$ (EigenGP performs similarly), both faster and more accurate.

\subsection{Prediction Quality on Nonstationary Data}
\begin{figure*}[t]
\begin{center}
\includegraphics[width=0.90\linewidth]{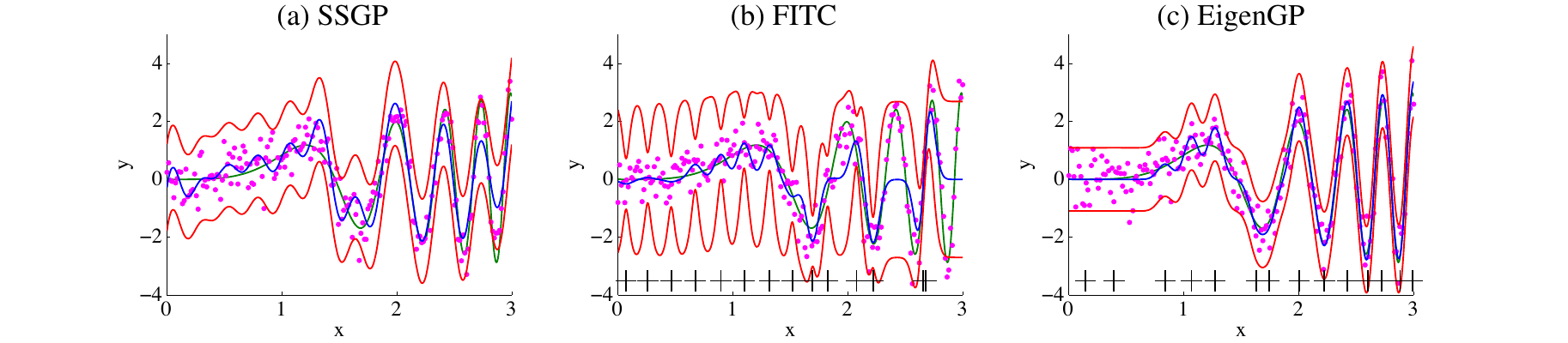}
\caption{Predictions on nonstationary data.
The pink dots correspond to noisy data around the true function $f(x)=x\sin(x^3)$, represented by the green curves. 
The blue and red solid curves correspond to 
the predictive means and $\pm$ two standard deviations around the means.
The black crosses near the bottom represent the estimated basis points for FITC and \ours.
} \label{fig:nonstationary}
\end{center}
\end{figure*}

We then compare all the sparse GP methods on an one-dimensional nonstationary synthetic dataset with 200 training and 500 test samples. The underlying  function is $f(x)=x\sin(x^3)$ where $x\in(0,3)$ and the standard deviation of the white noise is 0.5. This function is nonstationary in the sense that its frequency and amplitude increase when $x$ increases from 0.
We randomly generated the data 10 times and set the number of basis points (functions) to be 14 for all the competitive methods. 
Using the true function value as the label, we compute 
the means and the standard errors of NMSE and MNLP as in Table \ref{table:syn_nmse} (dataset 2). 
 For the Nystr\"{o}m method, 
the marginal likelihood optimization leads to much smaller error than the K-means based approach. However, both of them fare poorly when compared with the alternative methods. Table \ref{table:syn_nmse} also shows that
EigenGP and EigenGP$^*$ achieve a striking $\sim$25,000 fold error reduction compared with Nystr\"{o}m*, and a $\sim$10-fold error reduction compared with the second best method, SSGP. 
RVMs gave NMSE $0.0111\pm0.0004$ with $1.4\pm0.05$ seconds, averaged over 10 runs, while the results of EigenGP* with $M=50$ are NMSE $0.0110\pm0.0006$ with $0.89\pm0.1042$ seconds (EigenGP gives similar results).

We further illustrate the predictive mean and standard deviation on a typical run in Figure \ref{fig:nonstationary}. 
As shown in Figure \ref{fig:nonstationary}(a), the predictive mean of SSGP contains reasonable high frequency components for  $x\in(2,3)$ but, as a stationary GP model, these high frequency components give extra wavy patterns in the left region of $x$. In addition, the predictive mean on the right is smaller than the true one, probably affected by the small dynamic range of the data on the left. 
Figure \ref{fig:nonstationary}(b) shows that 
the predictive mean of FITC at $x\in(2,3)$ has lower frequency and smaller amplitude than the true function---perhaps influenced by the low-frequency part on the left $x\in(0,2)$. Actually because of the low-frequency part, FITC learns a large kernel width $\eta$; the average kernel width learned over the 10 runs is $207.75$. This  large value affects the quality of learned basis points (e.g., lacking of basis points for the high frequency region on the right). By contrast, using the same initial kernel width %0.44
as FITC, \egp learns a suitable kernel width---on average, $\eta = 0.07$---and provides good predictions as shown in Figure \ref{fig:nonstationary}(c).
%Finally we treated this dataset as a time series and compared the one-step-ahead prediction accuracy of these methods and popular time-series methods include AR, MA, ARIMA (see Appendix). Again, EigenGP performs best. 

\subsection{Accuracy vs. Time on Real Data}

\begin{figure}[th]
\begin{center}
\includegraphics[width=1\linewidth]{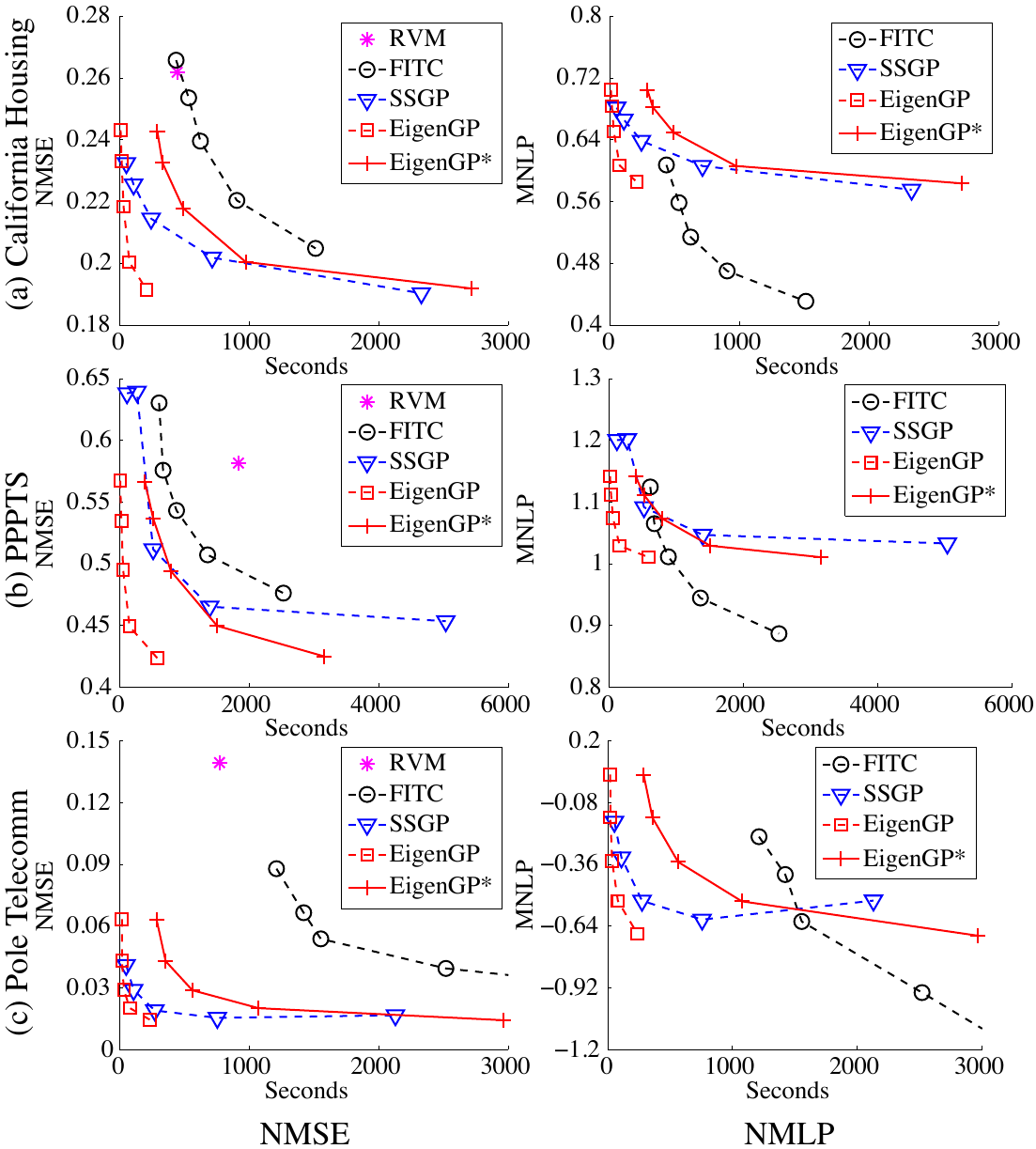}
\end{center}
\caption {NMSE and MNLP vs. training time. Each meth\-od (except RVMs) has five results associated with $M=25,50,100,200, 400$, respectively. In (c), the fifth result of FITC is out of the range; the actual training time is $3485$ seconds, the NMSE $0.033$, and the MNLP  $-1.27$. The values of  MNLP for RVMs are $1.30$, $1.25$ and $0.95$, %on the three datasets,
respectively.
 } \label{fig:real}
\end{figure}

To evaluate the trade-off between prediction accuracy and computational cost, we use three large real  datasets. 
The first dataset is California Housing \citep{Pace97sparsespacial}. 
We randomly split the 8 dimensional data into 10,000 training and 10,640 test  points. The second dataset is Physicochemical Properties of Protein Tertiary Structures (PPPTS) which can be obtained from \citet{Lichman:2013}. We randomly split the 9 dimensional data into 20,000 training  and 25,730 test points. The third dataset is Pole Telecomm that was used in \citet{Gredilla10}.  It contains 10,000 training and 5000 test samples, each of which has 26 features. We set %varied the rank or the number of basis functions (or points): 
$M=25, 50, 100, 200, 400$, and the maximum number of iterations in optimization to be 100 for all methods.

The NMSE, MNLP and the training time of these methods are shown in Figure \ref{fig:real}. 
In addition,  we ran the  Nystr\"{o}m method based on the marginal likelihood maximization, which is better than using K-means to set the basis points. 
Again, the  Nystr\"{o}m method performed orders of magnitude worse than the other methods: with $M=25, 50, 100, 200, 400$, on California Housing, the 
Nystr\"{o}m method uses $146$, $183$, $230$, $359$ and $751$ seconds for training, respectively, and gives the NMSE $917$, $120$, $103$, $317$, and $95$; on PPPTS, the training times are $258$, $304$, $415$, $853$ and $2001$ seconds and the NMSEs are $1.7\times10^5$, $6.4\times10^4$, $1.3\times10^4$, $8.1\times10^3$, and $8.1\times10^3$; and on Pole Telecomm, 
the training times are  $179$, $205$, $267$, $478$, and $959$ seconds and the NMSEs are $2.3\times10^4$, $5.8\times10^3$, $3.8\times10^3$, $4.5\times10^2$ and $84$.  The MNLPs are consistently large, and are omitted here for simplicity.

For RVMs, we include cross-validation in its training time because choosing an appropriate kernel width is crucial for RVM.  Since RVM learns the number of basis functions automatically from the data,  in Figure \ref{fig:real} it shows a single result for each dataset. 
\ours achieves the lowest prediction error using shorter time. Compared with \ours based on the sequential optimization, EigenGP$^*$ achieves similar errors, but takes longer because the joint optimization is more expensive.

\section{Conclusions}
In this paper we have presented a simple yet effective sparse Gaussian process method, \ours, and applied it to regression.  Despite its similarity to the Nystr\"{o}m method, \ours can improve its prediction quality by several orders of magnitude. 
\ours can be easily extended to conduct online learning by either using  stochastic gradient descent to update the weights of the eigenfunctions or applying the online VB idea for GPs \citep{HenFusLaw13}.

% \ours can also be applied to other learning tasks, including classification, semi-supervised learning and multi-task learning. 

\section*{Acknowledgments}
This work was supported by NSF ECCS-0941043, NSF CAREER award IIS-1054903, and the Center for Science of Information, an NSF Science and Technology Center, under grant agreement CCF-0939370.

\appendix
\bibliographystyle{named}
%\small
\bibliography{eigengp}

\begin{thebibliography}{}

\bibitem[\protect\citeauthoryear{Cressie and
  Johannesson}{2008}]{CressieAndJohannesson08}
Noel Cressie and Gardar Johannesson.
\newblock Fixed rank kriging for very large spatial data sets.
\newblock {\em Journal of the Royal Statistical Society: Series B (Statistical
  Methodology)}, 70(1):209--226, February 2008.

\bibitem[\protect\citeauthoryear{Csat\'{o} and Opper}{2002}]{Csato2002}
Lehel Csat\'{o} and Manfred Opper.
\newblock Sparse online {G}aussian processes.
\newblock {\em Neural Computation}, 14:641--668, March 2002.

\bibitem[\protect\citeauthoryear{de Leeuw}{2007}]{Leeuw07}
Jan de~Leeuw.
\newblock Derivatives of generalized eigen systems with applications.
\newblock In {\em Department of Statistics Papers}. Department of Statistics,
  UCLA, UCLA, 2007.

\bibitem[\protect\citeauthoryear{Faul and Tipping}{2001}]{Faul01analysisof}
Anita~C. Faul and Michael~E. Tipping.
\newblock Analysis of sparse {B}ayesian learning.
\newblock In {\em Advances in Neural Information Processing Systems 14}. MIT
  Press, 2001.

\bibitem[\protect\citeauthoryear{Hensman \bgroup \em et al.\egroup
  }{2013}]{HenFusLaw13}
James Hensman, Nicol{\`o} Fusi, and Neil~D. Lawrence.
\newblock Gaussian processes for big data.
\newblock In {\em the 29th Conference on Uncertainty in Artiﬁcial
  Intelligence (UAI 2013)}, pages 282--290, 2013.

\bibitem[\protect\citeauthoryear{Higdon}{2002}]{Higdon2002}
Dave Higdon.
\newblock Space and space-time modeling using process convolutions.
\newblock In Clive~W. Anderson, Vic Barnett, Philip~C. Chatwin, and Abdel~H.
  El-Shaarawi, editors, {\em Quantitative methods for current environmental
  issues}, pages 37--56. Springer Verlag, 2002.

\bibitem[\protect\citeauthoryear{Hoegaerts \bgroup \em et al.\egroup
  }{2005}]{Hoegaerts05}
Luc Hoegaerts, Johan A.~K. Suykens, Joos Vandewalle, and Bart~De Moor.
\newblock Subset based least squares subspace regression in {RKHS}.
\newblock {\em Neurocomputing}, 63:293--323, January 2005.

\bibitem[\protect\citeauthoryear{L\'{a}zaro-Gredilla \bgroup \em et al.\egroup
  }{2010}]{Gredilla10}
Miguel L\'{a}zaro-Gredilla, Joaquin Qui{\~{n}}onero-Candela, Carl~E. Rasmussen,
  and An\'{i}bal~R. Figueiras-Vidal.
\newblock Sparse spectrum {G}aussian process regression.
\newblock {\em Journal of Machine Learning Research}, 11:1865--1881, 2010.

\bibitem[\protect\citeauthoryear{Lichman}{2013}]{Lichman:2013}
Moshe Lichman.
\newblock {UCI} machine learning repository, 2013.

\bibitem[\protect\citeauthoryear{MacKay}{1992}]{MacKay92}
David~J.~C. MacKay.
\newblock {B}ayesian interpolation.
\newblock {\em Neural Computation}, 4(3):415--447, 1992.

\bibitem[\protect\citeauthoryear{Marzouk and Najm}{2009}]{Marzouk09}
Youssef~M. Marzouk and Habib~N. Najm.
\newblock Dimensionality reduction and polynomial chaos acceleration of
  {B}ayesian inference in inverse problems.
\newblock {\em Journal of Computational Physics}, 228(6):1862 -- 1902, 2009.

\bibitem[\protect\citeauthoryear{Minka}{2001}]{Minka01oldand}
Thomas~P. Minka.
\newblock Old and new matrix algebra useful for statistics.
\newblock Technical report, MIT Media Lab, 2001.

\bibitem[\protect\citeauthoryear{Pace and Barry}{1997}]{Pace97sparsespacial}
R.~Kelley Pace and Ronald Barry.
\newblock Sparse spatial autoregressions.
\newblock {\em Statistics and Probability Letters}, 33(3):291--297, 1997.

\bibitem[\protect\citeauthoryear{Paciorek and Schervish}{2004}]{PacSch03}
Christopher~J. Paciorek and Mark~J. Schervish.
\newblock Nonstationary covariance functions for {G}aussian process regression.
\newblock In {\em Advances in Neural Information Processing Systems 16}. MIT
  Press, 2004.

\bibitem[\protect\citeauthoryear{Qi \bgroup \em et al.\egroup
  }{2010}]{Qi:SparsePosteriorGaussianProc}
Yuan Qi, Ahmed~H. Abdel-Gawad, and Thomas~P. Minka.
\newblock Sparse-posterior {G}aussian processes for general likelihoods.
\newblock In {\em Proceedings of the 26th Conference on Uncertainty in
  Artificial Intelligence}, 2010.

\bibitem[\protect\citeauthoryear{Qui{\~{n}}onero-Candela and
  Rasmussen}{2005}]{Candela05Rasmussen}
Joaquin Qui{\~{n}}onero-Candela and Carl~E. Rasmussen.
\newblock A unifying view of sparse approximate {G}aussian process regression.
\newblock {\em Journal of Machine Learning Research}, 6:1935--1959, 12 2005.

\bibitem[\protect\citeauthoryear{Snelson and
  Ghahramani}{2006}]{Snelson06sparsegaussian}
Edward Snelson and Zoubin Ghahramani.
\newblock Sparse {G}aussian processes using pseudo-inputs.
\newblock In {\em Advances in Neural Information Processing Systems 18}. MIT
  press, 2006.

\bibitem[\protect\citeauthoryear{Tipping}{2000}]{Tipping00}
Michael~E. Tipping.
\newblock The relevance vector machine.
\newblock In {\em Advances in Neural Information Processing Systems 12}. MIT
  Press, 2000.

\bibitem[\protect\citeauthoryear{Titsias}{2009}]{Titsias09}
Michalis~K. Titsias.
\newblock Variational learning of inducing variables in sparse {G}aussian
  processes.
\newblock In {\em The 12th International Conference on Artificial Intelligence
  and Statistics}, pages 567--574, 2009.

\bibitem[\protect\citeauthoryear{Williams and Barber}{1998}]{WillamsBarber98}
Christopher~K.~I. Williams and David Barber.
\newblock {Bayesian classification with {G}aussian processes}.
\newblock {\em IEEE Transactions on Pattern Analysis and Machine Intelligence},
  20(12):1342--1351, 1998.

\bibitem[\protect\citeauthoryear{Williams and Seeger}{2001}]{Williams01}
Christopher~K.~I. Williams and Matthias Seeger.
\newblock Using the {N}ystr\"{o}m method to speed up kernel machines.
\newblock In {\em Advances in Neural Information Processing Systems 13},
  volume~13. MIT Press, 2001.

\bibitem[\protect\citeauthoryear{Yan and Qi}{2010}]{Yan10}
Feng Yan and Yuan Qi.
\newblock Sparse {G}aussian process regression via $\ell_1$ penalization.
\newblock In {\em Proceedings of 27th International Conference on Machine
  Learning}, pages 1183--1190, 2010.

\bibitem[\protect\citeauthoryear{Zhang \bgroup \em et al.\egroup
  }{2008}]{ZhangNystrom08}
Kai Zhang, Ivor~W. Tsang, and James~T. Kwok.
\newblock Improved {N}ystr\"{o}m low-rank approximation and error analysis.
\newblock In {\em Proceedings of the 25th International Conference on Machine
  Learning}, pages 1232--1239. ACM, 2008.

\end{thebibliography}

\end{document}